\documentclass[12pt,letterpaper]{article}

\usepackage{cmap}
\usepackage[T1]{fontenc}
\usepackage{geometry}
\usepackage{csquotes}
\usepackage{pslatex}
\usepackage{amsmath}
\usepackage{tipa}
\usepackage{tikz}
\usepackage{adjustbox}
\usepackage{booktabs}
\usepackage{microtype}
\PassOptionsToPackage{hyphens}{url}
\usepackage{hyperref}

\usepackage{fullpage}
\usepackage{setspace}

\usepackage{authblk}

\usepackage{todonotes}

\usepackage[
  backend=biber,
  style=apa,
  natbib=true,
  eprint=true,
  annotation=false,
]{biblatex}
\DeclareNameWrapperFormat{labelname:poss}{#1's}

\newrobustcmd*{\posscitealias}{%
  \AtNextCite{%
    \DeclareNameWrapperAlias{labelname}{labelname:poss}}}

\newrobustcmd*{\citeapos}{%
  \posscitealias
  \textcite}

\newrobustcmd*{\Posscite}{\bibsentence\posscite}

\newrobustcmd*{\posscites}{%
  \posscitealias
  \textcites}

\usepackage{float} 

\title{\Large{How communicatively optimal are exact numeral systems?\\Once more on lexicon size and morphosyntactic complexity}}

\author[1,*]{\mbox{Chundra Cathcart}}
\author[2]{\mbox{Arne Rubehn}}
\author[2]{\mbox{Katja Bocklage}}
\author[2]{\mbox{Luca Ciucci}}
\author[2]{\mbox{Kellen Parker van Dam}}
\author[2]{\mbox{Al\v{z}b\v{e}ta Ku\v{c}erov\'a}}
\author[2]{\mbox{Jekaterina Ma\v{z}ara}}
\author[2]{\mbox{Carlo Y. Meloni}}
\author[2]{\mbox{David Snee}}
\author[2]{\mbox{Johann-Mattis List}}
\affil[1]{Institute for the Interdisciplinary Study of Language Evolution

University of Zurich, Zurich, Switzerland}
\affil[2]{Chair for Multilingual Computational Linguistics

University of Passau, Passau, Germany}
\affil[*]{{\tt chundra.cathcart@uzh.ch}}

\date{February 19, 2026}

\begin{document}

\onehalfspacing

\maketitle

\begin{abstract}
\noindent Recent research argues that exact recursive numeral systems optimize communicative efficiency by balancing a tradeoff between the size of the numeral lexicon and the average morphosyntactic complexity (roughly length in morphemes) of numeral terms. 
We argue that previous studies have not characterized the data in a fashion that accounts for the degree of complexity languages display. 
Using data from 52 genetically diverse languages and an annotation scheme distinguishing between predictable and unpredictable allomorphy (formal variation), we show that many of the world's languages are decisively less efficient than one would expect. We discuss the implications of our findings for the study of numeral systems and linguistic evolution more generally.

\end{abstract}

\section{Introduction}

In the world's languages, numeral systems --- the inventories of terms used to denote different quantities --- show considerable diversity along a number of dimensions. 
Languages differ in terms of whether their numeral systems are exact, i.e., containing a unique term referring to every possible cardinality, or approximate, with one-to-many mappings between numeral terms and numerosities \citep{pica2004exact,nunez2017there,o2022cultural}. 
Different languages build their numeral systems around different arithmetic bases; while decimal systems are the most frequent cross-linguistically, vigesimal (base twenty) systems are also common and a variety of other bases are found as well \citep{comrie2022arithmetic}. 
Numerals also differ in terms of their morphosyntactic properties, exhibiting different orders of elements (i.e., modifiers with respect to bases) within numeral terms \citep{allassonniere2020numeral}. 
Much effort has been made to account for this variation. 
A popular line of research argues that numeral systems are subject to trade-offs between domains of complexity ultimately stemming from a pressure to facilitate efficient communication, similar to tradeoffs in other domains \citep[e.g.,][]{kemp2012kinship}. 

Here, we focus on another dimension of variation that has attracted some attention in recent years, namely that of formal variation within numeral systems \citep{koile2025decimal,cathcart2025complexity,rubehn2025annotating}. 
Specifically, we build on existing work which argues that languages optimize a tradeoff between the size of the lexicon used to construct numeral, comparing attested numeral systems to hypothetical artificially generated systems exhibiting a diverse range of configurations. 
We argue that previous studies have operationalized complexity in problematic ways and do not take into account a representative sample of languages exhibiting the full range of complexity found cross-linguistically. 
Improving upon these issues, we demonstrate that not all languages of the world exhibit optimality in the sense described by previous work. 
Our results serve as a reminder that putative universals rooted in communicative efficiency are not always as straightforward as they seem at first blush. 
We conclude with an appeal to take into account the full range of historical contingencies that shape linguistic evolution in at times unexpected ways.

\section{Background}

We take as a point of departure recent work \citep{denic2024recursive} which argues that exact numeral systems must manage a trade-off between (1) a pressure to keep the numeral lexicon as small as possible and (2) a need to keep numeral terms from 1 to 99 on average as morphosyntactically simple (i.e., involving the smallest combination of numeral terms and arithmetic operators) as possible. As an example of this dual pressure, languages like Japanese can generate numerals 1--99 with a small inventory of elements (the modifiers 1--9 and the base 10), but numbers above 20 are generated by concatenating together three elements (e.g., {\it ni j\={u} go} `25'), whereas a language with dedicated forms for each decade would have a larger inventory but would only need to combine two elements to express similar quantities (e.g., Swedish {\it tjugo fem} `25'). 
Drawing upon earlier work on communicative efficiency in numeral systems \citep{xu2020numeral}, the authors use a computational algorithm to generate a large number of hypothetical unattested numeral systems ranging from highly non-optimal (e.g., having a large numeric lexicon and high average morphosyntactic complexity) to optimal, and show that attested numeral systems lie along a Pareto frontier estimated from these synthetic data, meaning that they exhibit a near-optimal solution to the trade-off between the two variables. 

Subsequent work has modified aspects of this approach. \citet{yang2025re} note that the evolutionary algorithm used by \citet{denic2024recursive} to generate hypothetical numeral systems, based on \citeapos{hurford2011linguistic} grammar of numerals, lacks some properties of natural languages, including the use of suppletive forms 
(e.g., Ukrainian {\it sorok} `40', originally `bundle of [40] pelts', replacing earlier {\it \v{c}otyr-desjat} `$4 \times 10$'; \citealt{falowski2011east}). 
Altering the algorithm to incorporate this and other properties, the authors find that the original results hold. \citet{prasertsom2025recursive} call into question the metrics used in the original work, proposing alternative measures of complexity argued to better capture differences between attested and unattested systems (for the purpose of this paper we keep with the original study in investigating the trade-off between lexicon size and average morphosyntactic complexity, given the straightforwardness of operationalizing these properties of numeral systems under the approach we use). 

In this paper, we address and improve upon a major limitation of \citeapos{denic2024recursive} study involving the treatment of the data. 
As mentioned previously, one dimension of complexity which systems conceivably work to minimize is the size of the numeric lexicon. 
Curiously, the authors stress that they are concerned only (in their words) with the number of lexicalized numeral concepts found in a language, and not with allomorphic variation in form-meaning relationships (e.g., {\it ten} vs.\ {\it -teen} vs.\ {\it -ty}). 
In our view, this decision is not well motivated: at the very least, it strikes us as misguided to operationalize complexity whilst excluding information regarding unpredictable formal variation that learners of numeral systems must confront. 
After all, a central concern of most grammatical frameworks regardless of their theoretical orientation involves accounting for the exceptions that a language user must learn in order to produce and comprehend utterances.

Aside from these concerns, a more serious problem can be identified. 
As we will demonstrate, the premise that analyses can focus on lexicalized number meanings while ignoring form-meaning mappings is fundamentally flawed: any attempt to isolate lexicalization from the presence of irregular morphology is untenable, as the two phenomena are intertwined. 
It is not always possible to determine how \citeapos{denic2024recursive} coding scheme relates to the formal properties of the numerals under study, since they do not provide explicit information regarding how they arrived at their analyses. 
Nonetheless, a few examples will serve to illustrate the inconsistency that results from attempting to ignore irregular morphology. 

In many languages, the early teens are irregular exceptions within an otherwise transparent and predictable system. 
In the authors' coded data, English {\it eleven} is treated as a non-decomposable term lexicalizing the quantity $11$; however, German {\it elf} is represented as decomposed $10 + 1$. 
This analysis entails that {\it elf} contains allomorphs ({\it e-} and {\it -lf}?) of {\it ein(s)} `1' and {\it zehn} `10', even though the word contains no element formally resembling or cognate to the latter term, descending from Proto-Germanic {\it *aina-lifa-}, cf.\ Gothic {\it ain-lif} `11', lit.\ `one left (over)' \citep{kroonen2013etymological}. 
Persian {\it bist} `20' 
is analyzed as multimorphemic $2 \times 10$, despite not being straightforwardly decomposable into the numbers {\it du} `2' and {\it dah} `10'. Whether or not these are coding errors is beside the point: under an analysis that concerns itself with the lexicalization of number meanings but is not tasked with explaining the distribution of formal elements found in a numeral system (i.e., which allomorphs appear in which contexts), any analysis is valid for any number, as a concept can have any number of formal realizations. 
Concretely, there is nothing preventing us from analyzing {\it sorok} `40', arguably 
a simplex form, as a complex form (e.g., {\it sor-ok} `$4 \times 10$', with {\it sor-} and {\it -ok} treated as allomorphs of `4' and `10', respectively).\footnote{\citet{denic2024recursive} identify English, Persian, Russian, and other languages as unclear cases which are excluded from some analyses. This raises a further concern that certain results are biased in favor of simpler systems within their sample.} 
At worst, this analytical framework has the potential to collapse, treating all languages as identical in their numeral morphosyntax. At best, it leads to the inconsistencies we have 
highlighted 
above. 
This coding scheme likely captures some interesting variation stemming from the cross-linguistic use of different bases (where consistently coded) and arithmetic operators (e.g., subtraction), but remains far from characterizing the range of formal complexity that the world's languages display. 

We adopt a more principled approach, reconcilable with most morphological theories, which distinguishes between phonologically predictable and unpredictable (i.e., lexically specified) allomorphy resulting from sound change or suppletion (that is, the introduction of unrelated lexical material; \citealt{kim2016sound}). 
The former phenomenon consists of alternations such as that between Spanish {\it diec{\IPA [i]}siete} `17' and {\it diec{\IPA [j]}ocho} `18', where a segment surfaces as {\IPA [j]} before vowels and {\IPA [i]} elsewhere. 
An extreme example of phonologically unpredictable allomorphy is found in Dhivehi (dhiv1236, Maldives; \citealt{Fritz2002,GnanadesikanDhivehi2017}), where `7' has {\it s}- and {\it h}-initial allomorphs seen in {\it {sat\={a}}-ra} `17' and {\it {hat\={a}}-v\={\i}s} `27' (numerals are base-final). There is no straightforward phonological context that predicts when {\it s}- and {\it h}-initial variants surface; rather, speakers must remember to use the correct variant for the correct lexical item, just as Ukrainian speakers remember to use the suppletive form {\it sorok} `40' 
instead of a hypothetical form composed from the numerals $4$ and $10$. 
These unpredictable elements add to the amount of lexical material that is needed to characterize the entire system, and hence reflect the burden the system places on language users. 
We do not assume a holistic representation for irregulars, but allow for the possibility that variant forms are stored as sub-word units that can be abstracted from multiple irregular forms (e.g., English {\it thir-}, {\it fif-}, {\it -teen}, etc.). 

We work with phonemically transcribed, segmented data with manually annotated morphological glosses, making it straightforward to operationalize the morphological complexity of numerals in a language's system as well as the size of the system's inventory of lexical items. 
Hand-curated approaches of this sort often suffer from the problem of non-unique analyses \citep{round2017,wu_morphological_2019}; e.g., an annotator may choose an analysis {\it twelve} or {\it twe-lve}. 
We remedy this issue in several ways, including via inter-annotator reliability values and by 
showing that our results are robust to different approaches (permissive vs.\ strict) to morphological segmentability. 



\begin{table*}[]
    \centering
    \footnotesize{
    \begin{tabular}{llllll}
    \toprule
    Language & Concept & Source Form & Surface Form & Underlying Form & Morpheme Gloss\\
    \midrule
    French & {\sc ten} & {dix} & {\IPA dis} & {\IPA /dis/} & {\sc ten}\\
    French & {\sc sixteen} & {seize} & {\IPA sEz} & {\IPA /sEz/} & {\sc sixteen}\\
    French & {\sc seventeen} & {dix-sept} & {\IPA disEt} & {\IPA /dis-sEt/} & {\sc ten seven}\\
    French & {\sc eighteen} & {dix-huit} & {\IPA diz{\textturnh}it} & {\IPA /dis-{\textturnh}it/} & {\sc ten eight}\\
    \midrule
    Balochi & {\sc ten} & {d{\textschwa}} & {\IPA d@} & {\IPA /d@/} & {\sc ten}\\
    Balochi & {\sc sixteen} & {š\~azd{\textschwa}[g]} & {\IPA S\~azd@} & {\IPA /S\~az-d@/} & {\sc six$_2$ ten}\\
    Balochi & {\sc twenty} & {bist} & {\IPA bist} & {\IPA /bist/} & {\sc twenty}\\
    Balochi & {\sc twenty six} & {bist w š{\textschwa}šš} & {\IPA bistUS@S:} & {\IPA /bist-U-S@S:/} & {\sc twenty and six$_1$}\\
    \midrule
    Dhivehi & {\sc twelve} & {bāra} & {\IPA ba:ra} & {\IPA /ba:-ra/} & {\sc two$_2$ ten$_2$}\\
    Dhivehi & {\sc nineteen} & {onavihi} & {\IPA onaVihi} & {\IPA /ona-Vihi/} & {\sc minus\underline{\phantom{x}}one twenty$_1$}\\
    Dhivehi & {\sc twenty} & {vihi} & {\IPA Vihi} & {\IPA /Vihi/} & {\sc twenty$_1$}\\
    Dhivehi & {\sc twenty two} & {bāvīs} & {\IPA ba:Vi:s} & {\IPA /ba:-Vi:s/} & {\sc two$_2$ twenty$_2$}\\
    \bottomrule
    \end{tabular}
    }
    \caption{Selected examples of annotation scheme from French (stan1290; \citealt{French}), Balochi (west2368; \citealt{BarkerMengal1969}) and Dhivehi (dhiv1236; \citealt{GnanadesikanDhivehi2017}), slightly modified for increased clarity. Phonologically predictable allomorphs (e.g., French {\IPA [di]} before {\IPA /s/} vs.\ {\IPA [diz]} before a voiced segment vs.\ {\IPA [dis]} elsewhere) are coded as underlyingly the same, whereas allomorphs with no clear phonological conditioning environment (e.g., Balochi {\IPA [S@S:]} vs.\ {\IPA [S\~az]} `6', Dhivehi {\IPA [Vihi]} vs.\ {\IPA [Vi:s]} `20') are coded as separate underlying morphemes. We exclude cognate codings, as cognacy is not relevant to the research question investigated here, which depends only on the synchronic configurations exhibited by languages.}
    \label{tab:exx}
\end{table*}

\section{Materials and methods}

\subsection{Manually segmented and glossed data}

We employ a revised and enlarged collection of manually segmented and glossed numerals, building upon the ``Compositional Structures in Numeral Systems'' (CoSiNuS, \citealt{rubehn2025annotating}) database and extending the sample to include the forms for numbers 1 to 99 in 52 genetically and geographically diverse languages. Word forms are transcribed in a unified transcription system with words segmented into morphemes, which are assigned to language-internal cognate sets and annotated with morpheme glosses \citep{Hill2017a}, as in Table \ref{tab:exx}.  
The morpheme annotation scheme distinguishes between different unpredictable morphemic variants pertaining to the same meaning; e.g., English {\it five} ({\sc five$_1$}) and {\it ten} ({\sc ten$_1$}) vs.\ {\it fif-teen} ({\sc five$_2$ ten$_2$}) and {\it fif-ty} ({\sc five$_2$ ten$_3$}). 
The annotation furthermore distinguishes between phonologically unpredictable and predictable allomorphy by checking the literature for phonological rules that could generate variant surface forms from the same underlying representation, explicitly documenting potentially controversial decisions. 
All analyses were checked and critiqued by multiple coders. 
We calculated the inter-annotator agreement scores for both a $10$\% sample of all data points and for the complete lists of numerals for 10 randomly sampled languages. Agreement between annotators was $87$\% overall ($N=529$) and $94.5$\% for the complete datasets of 10 languages ($N=1003$), indicating consistency of the segmentation scheme. 
Data are stored in in Cross-Linguistic Data Formats \citep{Forkel2018a}.

Some cases in which annotator-level idiosyncrasies may impact results warrant discussion. 
For example, in situations where we are forced to infer the operation of productive synchronic processes (e.g., assimilation, voicing, vowel raising, reduction, etc.), our analyses may underestimate the number of unpredictable allomorphs in a given system. In general, our analyses tend to lump rather than split; we treat German {\it -zig} and {\it -ßig} (in \textit{zwan{\IPA [ts]}ig} `20', \textit{drei{\IPA [s]}ig} `30', \textit{vier{\IPA [ts]}ig} `40', etc.) as underlyingly the same, since the latter occurs only after a vowel, despite the lack of secure evidence for similar alternations elsewhere in the lexicon. However, as we discuss later, this does not harm the overall thrust of (and in fact further strengthens) the argument we make.  
Another issue is the following: while we have taken pains to be consistent in only treating recurrent phonological elements as segmentable morphemes, it could be argued that we over-segment in some cases, inflating morphosyntactic complexity. 
As an example, we segment the early teens in Dhivehi as {\it eg\=a-ra} `11', {\it b\=a-ra} `12', {\it t\=e-ra} `13', etc.\ on the basis of recurrent {\it -ra} ({\sc ten$_2$}); this creates a unique morpheme {\it eg\=a-} found only in this context ({\it b\=a-} and {\it t\=e-} appear elsewhere). 
It could be argued that {\it eg\=ara} should not be segmented. 
For robustness, we recode the data such that segmented forms where at least one morpheme is found only once are treated as holistic, non-decomposed irregular forms. 
We present results of analyses carried out using this recoded data with its {\sc narrow} construal of segmentable allomorphy in addition to the original {\sc broad} representation.





\subsection{Lexicon size and morphosyntactic complexity}

We closely follow the studies cited above in calculating lexicon size and morphosyntactic complexity for the broad and narrow representations of our data. We calculate lexicon size by counting up the number of unique unpredictable allomorphs in a numeral system. 
Morphosyntactic complexity is a weighted average for numerals 1 to 99 defined as proportional to $\sum_{i=1}^{99} |w_i|\cdot i^{-2}$, where $|w_i|$, the length of the term denoting cardinality $i$, is defined as the number of morphemes and arithmetic operators (covert and overt) that it contains (i.e., German {\it ein-und-zwan-$\emptyset$-zig} `1 + 2 $\times$ 10' and English {\it twen-$\emptyset$-ty-$\emptyset$-one} `2 $\times$ 10 + 1' have the same length), and $i^{-2}$ is proportional to the approximate frequency of cardinality $i$ in usage. 
We exclude forms not pertaining to numeral meaning or arithmetic operations (e.g., morphemes involved in subtraction) from the calculation of lexicon size and complexity, as they generally reflect language-specific artifacts that have nothing to do with numeral complexity (e.g., citation forms for numerals in Georgian contain a case suffix which should not count toward the size of the numeral lexicon or the number of morphemes in a numeral term). 
In cases where languages contain multiple terms for the same cardinality, our annotation schema marks a default form, which we use in calculating these quantitites. 

%
%





\begin{figure*}[t]

    \centering
    \adjustbox{max width=.96\linewidth}{
    \input{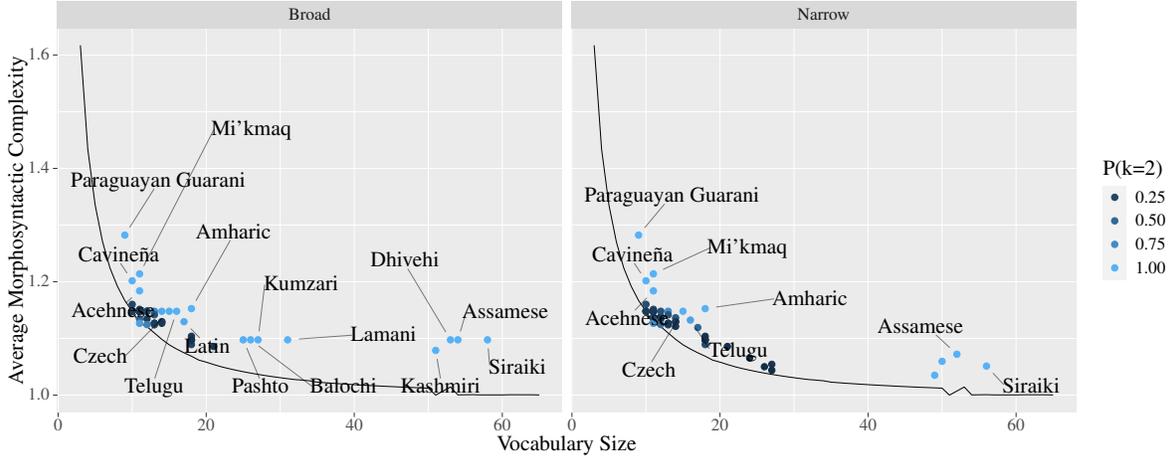}
    }

    \caption{Lexicon sizes plotted against morphosyntactic complexity values for languages in our sample, under broad and narrow data formats, with selected languages labeled. Colors represent posterior probabilities of component membership (lighter values indicate higher $P(k=2)$). The solid line represents the Pareto frontier estimated by the evolutionary algorithm.}
    \label{fig:complexity_plot}
\end{figure*}

\subsection{Synthetic numeral systems}

We generate hypothetical non-optimal and optimal numeral systems using \citeapos{yang2025re} implementation of the evolutionary algorithm originally proposed by \citet{denic2024recursive}. The algorithm generates hypothetical numeral systems using \citeapos{hurford2011linguistic} grammar of numeral phrases, randomly sampling candidate bases, digits, and irregular suppletive forms to denote numerals. We modify this implementation to allow up to 98 bases, digits, and suppletive forms in a given system. 
We evolve 100 hypothetical systems for 300 generations, randomly permuting the numeral grammar and selecting more optimal (i.e., reducing complexity for one variable without increasing complexity for the other) outcomes of this permutation, the end result of which consists of 100 hypothetical numeral systems lying along the Pareto frontier representing the optimal tradeoff between lexicon size and complexity. For ease of visualization, we compare attested systems only to the 
optimal systems populating this frontier. 

Note that the representation of numeral systems are in some ways not directly comparable to the attested data that we analyze; while hypothetical systems contain suppletive forms replacing a numeral expression with a single non-compositional form (e.g., German {\it elf} `11', Kashmiri {\it k{\textschwa}h} `11', Persian {\it bist} `20'), this process does not simulate unpredictable allomorphy at the sub-word level (e.g., {\it thir-teen}). Nonetheless, the optimal languages generated serve as a valid baseline against which to compare attested systems under our analysis, as they indicate, given a numeral lexicon size, how a highly economical language would distribute numeral forms across expressions.

\subsection{Mixture modeling}

We employ Bayesian mixture modeling to infer whether the data in our sample can be partitioned into separate components exhibiting different types of behavior. Our rationale is that some languages in our sample may cluster into different groups characterized by more vs.\ less optimal relationships between lexicon size and average morphosyntactic complexity. 
We assume a mixture of regression models with the following generative story: for each language $\ell = 1..\mathcal{L}$, a component label is generated $z_\ell \sim \text{Categorical}(\boldsymbol{\pi})$, 
where $\pi_k = K^{-1}:k=1..K$ ($K$ denotes the number of mixture components; we fixed this parameter to mitigate identifiability and convergence issues); 
the average morphosyntactic complexity value for a given language is generated on the basis of the sampled component label and the language's lexicon size, $\text{\tt morphosyntactic\underline{\phantom{x}}complexity}_\ell | z_\ell = k \sim \text{LogNormal}(\alpha_k + \beta_k \cdot \text{\tt lexicon\underline{\phantom{x}}size}_\ell,\sigma_k)$, with $\text{Normal}(0,1)$ priors over $\boldsymbol{\alpha}$ and $\boldsymbol{\beta}$ and a $\text{HalfNormal}(0,1)$ prior over $\boldsymbol{\sigma}$. 
During inference we marginalize out the discrete parameter $\boldsymbol{z}$, and reconstruct it from the fitted posterior samples of continuous parameters. 
We use RStan \citep{Carpenteretal2017} to fit mixture models to the values for these variables found in the attested languages in our sample for $K = 1..4$ over four chains of 2000 iterations of the No U-Turn Sampler (NUTS), discarding the first half of samples and assessing convergence via standard diagnostics \citep[e.g.,][]{GelmanRubin1992}. 
We carry out model comparison via leave-one-out expected log pointwise density (ELPD) values \citep{vehtari2017practical} and model stacking \citep{loo2017}, which averages predictive distributions of different models to generate weights representing their relative predictive power. 
In general, models with $K > 2$ did not converge, so we compare a two-component mixture model to a null one-component model (the equivalent of a standard regression model). 
The two-component mixture model is a better fit to the data, with a higher expected log pointwise density 
(broad: $\text{ELPD}_{K=2} = 122.2, \text{ELPD}_{K=1} = 107.8, \Delta_{\text{ELPD}} = -14.4, SE_{\Delta} = 6.6$; 
narrow: $\text{ELPD}_{K=2} = 132.8, \text{ELPD}_{K=1} = 104.2, \Delta_{\text{ELPD}} = -28.6, SE_{\Delta} = 6.3$) 
and stacking weight (broad: $0.870$ vs.\ $0.130$; narrow: $1.0$ vs.\ $0.0$). 
We classify languages according to the posterior probability of their component membership. 

\section{Results}


Figure \ref{fig:complexity_plot} plots lexicon sizes against average morphosyntactic complexity values for attested languages in our sample, under broad and narrow coding schemes. The Pareto frontier representing optimal tradeoffs between lexicon size and complexity estimated by the evolutionary algorithm is shown by the black curve along the lower left side of the plot. 
As is typically the case in studies of this sort, most languages lie slightly above the Pareto frontier rather than directly on the curve itself. 
This configuration represents a near-optimal solution to the problem of minimizing both lexicon size and complexity. 
At the same time, a number of points lie relatively far from the curve. 
To our knowledge, despite a large body of literature comparing near-optimal systems to optimal and non-optimal ones, there are no carefully developed criteria for determining how far away from the frontier an attested language must lie in order to no longer be considered near optimal. 
Furthermore the magnitude of these deviations is small and difficult to interpret in meaningful terms. 

We turn to the results of our mixture model in order to assess whether there are statistically detectable differences in the behavior of different languages that can be interpreted in a straightforward fashion. 
The points representing joint values of the variables of interest in attested languages are colored according to their posterior probability of mixture component membership. 
Visually, there is a clear separation between data points assigned to component $k=1$ (dark points, tightly straddling the frontier) and $k=2$ (light points, further from the frontier). 
The parameters of the different mixture components are even more telling. The mean slope of the lognormal-link regression function associated with component $k=1$ is negative 
(broad: $\hat{\beta_1}=-0.005$, 95\% percentile interval $=[-0.006,-0.004]$; 
narrow: $\hat{\beta_1}=-0.005$, 95\% PI $=[-0.005,-0.005]$), while the posterior distribution of the slope for component $k=2$ is effectively (or at least closer to) zero 
(broad: $\hat{\beta_2}=-0.001$, 95\% PI $=[-0.002,-0.0007]$; 
narrow: $\hat{\beta_2}=-0.002$, 95\% PI $=[-0.003,-0.001]$). 
What this indicates is that for languages associated with component $k=2$, average morphosyntactic complexity is not expected to decrease (or increase) as lexicon size increases.

This is borne out visually when we inspect the behavior of this component: when we look at languages with a lexicon size of around 30 or greater, they seem to follow a straight line with a slope of zero. In this group of languages, no matter how large their inventory, they do not decrease in complexity. 
These are languages in our sample (primarily Indo-European and Semitic) containing irregular forms, either in the teens (e.g., English {\it eleven}, Spanish {\it once}, Persian {\it yaz-dah}) or beyond, all of which increase the lexicon size of the system without adding to the number of usable bases and thus reducing morphosyntactic complexity. 
Outlying Indo-Aryan languages (Lamani, Kashmiri, Dhivehi, Siraiki) tolerate a high degree of irregularity as well. 
Languages like Paraguayan Guaraní exhibit higher complexity than expected given its lexicon size because of its hybrid use of different bases. 
The languages in $k=1$ (dark points in a straight line showing a clearer negative relationship between the two variables of interest) consist of languages from other families with highly transparent systems and different bases (quinary, decimal, vigesimal, etc.), where the tradeoff between the two variables is far more linear. 

\section{Discussion}

Our results show that while a number of languages present a near-optimal solution to the pressure to minimize both lexicon size 
and average morphosyntactic complexity, 
many languages do not, lying relatively far from the Pareto frontier and forming a distinct, statistically detectable group to the exclusion of languages with more efficient systems. 
\citet{denic2024recursive} argue that while all languages' numeral systems, exact and approximate, must optimize a tradeoff between the number of distinct numerosities they can express and the informativity of the system \citep{xu2020numeral}, exact systems must manage an orthogonal tradeoff between lexicon size and morphosyntactic complexity. 
We have shown that while many exact numeral systems manage this tradeoff efficiently, a considerable number do not; this result holds despite the potential underestimation of irregularity on our part (e.g., German {\it -zig} vs.\ {\it -ßig}) and under two data coding schemes. 
The key dimension underlying this difference 
is regularity: 
when taking into account the number of unpredictable elements in languages' numeral term inventories --- and we have argued above that it is not possible to operationalize lexicon size without taking allomorphy into account --- it turns out that certain languages tolerate a varying degree of irregularity in numeral systems with no apparent benefit (in terms of the variables discussed above). 
At the extreme end of this tolerance for irregularity are Indo-Aryan numeral systems; though long recognized for their unusual degree of irregularity \citep{bright1972,comrie2011,hurford1987language,hammarstrom2010rarities,Overmann2025}, these languages do not figure prominently in quantitative studies of complexity in numeral systems. 
The Indo-Aryan languages make up only a small, phylogenetically restricted fraction of the 7000-odd languages spoken today, yet they are spoken by a large proportion of the world's population. 
Although the history of these systems is relatively well documented and the historical developments affecting them are reasonably well understood \citep{berger1992modern,andrijanic2024hindi}, the reasons for the development of such a high degree of complexity remain unclear, as well as the mechanisms involved in its persistence for the last 1000 years or so. 
It is possible that a mechanistically grounded reason for this irregularity can be found, given the irregularity is argued to have several communicative benefits, such as earlier cues to form identity that allow listeners to process forms more quickly \citep{Blevinsetal2017}. It remains to be seen whether the preservation of irregularity in these systems serves to enhance the discriminability of individual numeral forms.

Given that the irregularity exhibited by Indo-Aryan numeral systems breaks many generalizations regarding communicative efficiency, what are some properties that all numeral systems have in common? 
It is widely observed that morphological irregularity is more likely to emerge and persist in highly frequent words than in less frequent ones \citep{sims-williams_token_2021}. 
This generalization also holds for numeral systems like that of English, where irregular suppletive forms like {\it twelve} have lower cardinality (and therefore higher frequency; \citealt{dehaene1992cross}) than regular, decomposable forms like {\it ninety-two} \citep{Brysbaert2005}. 
Indo-Aryan numeral systems are different from those of English, Persian, and Neo-Aramaic in that there is no clear binary opposition between regularity and irregularity: irregular-looking forms can be found beyond the teens, all the way up to 99, and numerals seem to occupy a cline spanning from somewhat irregular to highly irregular. 
Nevertheless, it has been shown that this generalization holds for Indo-Aryan numeral systems as well: 
despite their higher overall irregularity in comparison to other systems, higher-cardinality (lower-frequency) numerals are more predictable (according to a variety of computational models) and re-use more recurrent elements than lower-cardinality ones \citep{cathcart2017decomposability,cathcart2025complexity}. 
Impressionistically, this makes sense when inspecting forms in selected Indo-Aryan languages: for example, forms in the Dhivehi twenties display ample allomorphic variation (e.g., {\itshape vihi} `twenty', {\itshape ek\=a-v\={\i}s} `twenty one', {\itshape sab-b\={\i}s} `twenty six') in comparison to the seventies (e.g., {\itshape ha$\mathring{t}$tari} `seventy', {\it ek\=a-ha$\mathring{t}$tari} `seventy one', {\itshape sa-ha$\mathring{t}$tari} `seventy six'). This intuition has robust statistical support. 

This issue is related to some of the pressures involved in minimizing morphosyntactic complexity discussed by \citet{denic2024recursive}. 
The average morphosyntactic complexity measure proposed by the authors and used in this paper, though an aggregate measure, 
reflects the tendency 
to keep more frequent elements at lower cardinalities shorter (in terms of the number of elements combined) than less frequent elements at higher cardinalities. 
The authors link this pressure to a number of general principles in language use, invoking Zipf's law of abbreviation \citep{Zipf1949} --- the tendency of more frequent elements to be shorter --- as well as the idea that more frequently used elements are accessed whole as opposed to stored in a decomposed representation. 
We generally concur with these ideas: it seems to be the case that less frequent numerals of higher cardinality are phonologically longer, more transparent, and more decomposable than more frequent ones. This can be operationalized in a variety of ways, including those that do not necessitate the manual segmentation of words into morphemes \citep{baayen2018inflectional,beniamine2021multiple,guzman2024analogical}, as we have done here, and future research can aid in understanding the evolutionary mechanisms that maintain this state of affairs. 
However, as we have shown here, the drive to keep more frequent numerals phonologically shorter and more transparent does not necessarily correspond to differences in the number of segmentable morphological units in a given word. Our analyses find quasi-recurrent elements (and evidence of multimorphemicity) even in the highly irregular teens. 
Furthermore, we show that the pressure to minimize the average morphosyntactic complexity of utterances does not interact with the pressure to minimize complexity of the numeral lexicon in a meaningful way across languages: there exist a number of languages with a large inventory of terms that do not distribute them in a way that minimizes length in morphemes. Ultimately, we are likely observing not only the effects of cognitive pressure but of social factors as well (cultural diffusion, trade networks, contact resulting in hybrid systems). Our understanding of pressures toward efficiency should accommodate these local historical contingencies.











\section{Conclusion}
This study questions the notion that all numeral systems provide a near-optimal solution to the twin pressures of lexicon size and morphosyntactic complexity minimization. 
Our research departs from previous work by taking into account the full range of morphological complexity these systems display, which we argue cannot be ignored. 
Our results show that irregularity complicates the picture, particularly the high degree of irregularity displayed by Indo-Aryan languages.

While our results challenge the lexicon size/complexity trade-off from a synchronic perspective --- there exist contemporary languages that fly in the face of this generalization --- they do not rule out the possibility that these pressures exert themselves in the context of diachronic linguistic evolution. 
Many Indo-Aryan languages have been highly irregular for at least 1000 years, but what is to say that they won't evolve to balance these pressures in the next 1000 years? 
Further work is needed to elucidate the dynamics of these pressures in language change, and to better understand the origins and life cycle of deviations from efficiency.

\section*{Supplementary Materials}

The data underlying our analyses are curated on GitHub under \url{https://github.com/numeralbank/cosinus} (v2.0) and archived on Zenodo under \url{https://doi.org/10.5281/zenodo.18684984}. 
The code used in our analyses 
is curated on GitHub under \url{https://github.com/chundrac/numeral-complexity-tradeoff} and archived on Zenodo under \url{https://doi.org/10.5281/zenodo.18696081}. 

\section*{Acknowledgments}

We are grateful to David Yang for guidance on modifying his code. 
C.C. was supported by the NCCR Evolving Language (SNSF Agreement No. 51NF40\underline{\phantom{X}}180888), and gratefully acknowledges Swiss National Science Foundation grant No.\ 207573. A.R., K.B., L.C., A.K., C.Y.M., D.S., and J.-M.L. were supported by the ERC Consolidator Grant ProduSemy (PI J.-M.L., Grant No. 101044282, see \url{https://doi.org/10.3030/101044282}). Views and opinions expressed are however those of the author(s) only and do not necessarily reflect those of the European Union or the European Research Council Executive Agency (nor any other funding agencies involved). Neither the European Union nor the granting authority can be held responsible for them.

\section*{Author contributions}
\begin{description}
\item C.C.: Conceptualization, Methodology, Software, Visualization, Data curation, Writing (Original Draft)
\item A.R.: Project administration, Data curation, Writing (Review \& Editing)
\item J.-M.L.: Project administration, Data curation, Writing (Review \& Editing)
\item K.B., L.C., K.P.D., A.K., C.Y.M., D.S.: Data curation
\item J.M.: Validation
\end{description}

\printbibliography

@article{falowski2011east,
  title={The {E}ast-{S}lavonic sorok `40’ revisited},
  author={Fa{\l}owski, Adam},
  journal={Studia Etymologica Cracoviensia},
  volume={16},
  number={1},
  pages={7--15},
  year={2011},
  publisher={Wydawnictwo Uniwersytetu Jagiello{\'n}skiego}
}

@Article{Hill2017a,
  author      = {Hill, Nathan W. and List, Johann-Mattis},
  journal     = {Yearbook of the Poznań Linguistic Meeting},
  title       = {{C}hallenges of annotation and analysis in computer-assisted language comparison: {A} case study on {B}urmish languages},
  year        = {2017},
  doi         = {https://doi.org/10.1515/yplm-2017-0003},
  number      = {1},
  pages       = {47–76},
  url         = {https://www.degruyter.com/view/j/yplm.2017.3.issue-1/yplm-2017-0003/yplm-2017-0003.xml},
  volume      = {3},
  _code       = {https://github.com/digling/challenges-of-annotation-paper},
  _pdf        = {https://www.degruyter.com/downloadpdf/j/yplm.2017.3.issue-1/yplm-2017-0003/yplm-2017-0003.xml},
  _supplement = {https://zenodo.org/badge/latestdoi/94350448},
  groups      = {Papers},
  keywords    = {_usesLingPy, computer-assisted approach, Burmish languages, annotation, _calc},
  sortauthor  = {List, 9},
  timestamp   = {2019-08-04},
}

@article{kim2016sound,
  title={From sound change to suppletion: case studies from {Indo-European} languages},
  author={Kim, Ronald I},
  journal={Slovo a slovesnost},
  volume={77},
  number={4},
  pages={354--370},
  year={2016},
  publisher={AV {\v{C}}R-Akademie v{\v{e}}d {\v{C}}esk{\'e} republiky-{\'U}stav pro jazyk {\v{c}}esk{\`y}}
}

@article{Carpenteretal2017,
	Author = {Carpenter, Bob and Gelman, Andrew and Hoffman, Matthew D. and Lee, Daniel and Goodrich, Ben and Betancourt, Michael and Brubaker, Marcus and Guo, Jiqiang and Li, Peter and Riddell, Allen},
	Journal = {Journal of statistical software},
	Number = {1},
	Pages = {1-32},
	Title = {{Stan: A probabilistic programming language}},
	Volume = {76(1)},
	Year = {2017}}

@article{GelmanRubin1992,
	Author = {Andrew Gelman and Donald B. Rubin},
	Journal = {Statistical Science},
	Pages = {457-511},
	Title = {{Inference from iterative simulation using multiple sequences}},
	Volume = {7},
	Year = {1992}}

@article{loo2017,
title = {Using stacking to average {Bayesian} predictive distributions},
author = {Yuling Yao and Aki Vehtari and Daniel Simpson and Andrew Gelman},
year = {2017},
journal = {Bayesian Analysis},
doi = {10.1214/17-BA1091},
}

@article{vehtari2017practical,
	Author = {Vehtari, Aki and Gelman, Andrew and Gabry, Jonah},
	Journal = {Statistics and computing},
	Number = {5},
	Pages = {1413--1432},
	Publisher = {Springer},
	Title = {{Practical {Bayesian} model evaluation using leave-one-out cross-validation and WAIC}},
	Volume = {27},
	Year = {2017}}

@inproceedings{rubehn2025annotating,
  title={Annotating and inferring compositional structures in numeral systems across languages},
  author={Rubehn, Arne and Rzymski, Christoph and Ciucci, Luca and Bocklage, Katja and Ku{\v{c}}erov{\'a}, Al{\v{z}}b{\v{e}}ta and Snee, David and Stephen, Abishek and Van Dam, Kellen Parker and List, Johann-Mattis},
  booktitle={Proceedings of the 7th Workshop on Research in Computational Linguistic Typology and Multilingual NLP},
  pages={29--42},
  year={2025}
}

@incollection{hammarstrom2010rarities,
  title={Rarities in numeral systems},
  author={Hammarstr{\"o}m, Harald},
  booktitle={Rethinking universals: How rarities affect linguistic theory},
  editor={Jan Wohlgemuth and Michael Cysouw},
  pages={11--60},
  year={2010},
  address={Berlin},
  publisher={De Gruyter Mouton}
}

@article{koile2025decimal,
  title={Decimal systems around the world},
  author={Koile, Ezequiel and Blasi, Dami{\'a}n},
  journal={Philosophical Transactions of the Royal Society B: Biological Sciences},
  volume={380},
  number={1937},
  year={2025},
  publisher={The Royal Society}
}

@article{cathcart2025complexity,
  title={Complexity counts: global and local perspectives on Indo-Aryan numeral systems},
  author={Cathcart, Chundra},
  journal={arXiv preprint arXiv:2505.21510},
  year={2025}
}

@article{wu_morphological_2019,
	title = {Morphological Irregularity Correlates with Frequency},
	url = {http://arxiv.org/abs/1906.11483},
	author = {Wu, Shijie and Cotterell, Ryan and O'Donnell, Timothy J.},
	year = {2019},
    journal={arXiv preprint arXiv:1906.11483}
}

@book{kroonen2013etymological,
  title={Etymological Dictionary of {Proto-Germanic}},
  author={Kroonen, Gus},
  year={2013},
  publisher={Brill}
}

@article{prasertsom2025recursive,
  title={Recursive numeral systems are highly regular and easy to process},
  author={Prasertsom, Ponrawee and Silvi, Andrea and Culbertson, Jennifer and Johansson, Moa and Dubhashi, Devdatt and Smith, Kenny},
  journal={arXiv preprint arXiv:2510.27049},
  year={2025}
}

@article{round2017,
url = {https://doi.org/10.1515/flin-2017-0027},
title = {{Matthew K. Gordon}: {P}honological typology},
author = {Erich R. Round},
pages = {745--755},
volume = {51},
number = {3},
journal = {Folia Linguistica},
doi = {doi:10.1515/flin-2017-0027},
year = {2017},
lastchecked = {2025-09-10}
}

@inproceedings{yang2025re,
  title={Re-examining the tradeoff between lexicon size and average morphosyntactic complexity in recursive numeral systems},
  author={Yang, David and Regier, Terry},
  booktitle={Proceedings of the 47th Annual Conference of the Cognitive Science},
  editors={D. Barner and N.R. Bramley and A. Ruggeri and C.M. Walker},
  pages={863--869},
  year={2025}
}

@book{hurford2011linguistic,
  title={The linguistic theory of numerals},
  author={Hurford, James R},
  address={Cambridge},
  year={1975},
  publisher={Cambridge University Press}
}

@article{denic2024recursive,
  title={Recursive Numeral Systems Optimize the Trade-off Between Lexicon Size and Average Morphosyntactic Complexity},
  author={Deni{\'c}, Milica and Szymanik, Jakub},
  journal={Cognitive Science},
  volume={48},
  number={3},
  pages={e13424},
  year={2024},
  publisher={Wiley Online Library}
}

@article{guzman2024analogical,
  title={An analogical approach to the typology of inflectional complexity},
  author={Guzm{\'a}n Naranjo, Mat{\'\i}as},
  journal={Journal of Language Modelling},
  volume={12},
  year={2024}
}

@incollection{bright1972,
    author = {William Bright},
    title = {Hindi numerals},
    booktitle = {Studies in linguistics in honor of {George L.\ Trager}},
    publisher = {Mouton},
    address = {The Hague},
    editor = {M. Estellie Smith},
    pages = {222-230},
    year = {1972}
}

@incollection{andrijanic2024hindi,
  title={Hindi Cardinal Numerals in a Historical and Comparative Perspective},
  author={Andrijani{\'c}, Ivan},
  editor={Andrijani{\'c}, Ivan and Monika Browarczyk},
  address={Zagreb},
  publisher={FF Press},
  booktitle={Between Language and Literature: Hindī in Classroom and Beyond},
  pages={151--170},
  year={2024}
}

@Inbook{Overmann2025,
author="Overmann, Karenleigh A.",
title="Numbers in {I}ndia",
bookTitle="Cultural Number Systems: A Sourcebook",
year="2025",
publisher="Springer Nature Switzerland",
address="Cham",
pages="163--169",
url="https://doi.org/10.1007/978-3-031-83383-0_25"
}

@misc{French,
    author = {Doherty, Liam},
    title = {wikidict-fr: {Wikipedia Bilingual Reference Data (French)}},
    url = {https://github.com/open-dict-data/wikidict-fr},
    year = {2016}
}

@misc{comrie2011,
  title={Typology of numeral systems},
  author={Bernard Comrie},
  date={2011},
  url={https://lingweb.eva.mpg.de/channumerals/TypNum_Latest_21ho.pdf}
}

@book{BarkerMengal1969,
    Address = {Montreal},
    Author = {Barker, Muhammad Abd-al-Rahman and Mengal, Aqil Khan},
    Publisher = {Institute of Islamic Studies, McGill University},
    Title = {A Course in {B}aluchi},
    Year = {1969}
}

@article{kemp2012kinship,
  title={Kinship categories across languages reflect general communicative principles},
  author={Kemp, Charles and Regier, Terry},
  journal={Science},
  volume={336},
  number={6084},
  pages={1049--1054},
  year={2012},
  publisher={American Association for the Advancement of Science}
}

@book{Zipf1949,
title={Human behavior and the principle of least effort},
year={1949},
publisher={Addison-Wesley Press},
address={Cambridge, MA},
author={Zipf, George Kingsley}
}

@incollection{berger1992modern,
  title={Modern {Indo-Aryan}},
  author={Berger, Hermann},
  booktitle={{Indo-European} numerals},
  editor={Jadranka Gvozdanovi\'{c}},
  address={Berlin, New York},
  publisher={Mouton de Gruyter},
  pages={243--287},
  year={1992}
}

@article{xu2020numeral,
  title={Numeral systems across languages support efficient communication: From approximate numerosity to recursion},
  author={Xu, Yang and Liu, Emmy and Regier, Terry},
  journal={Open Mind},
  volume={4},
  pages={57--70},
  year={2020},
  publisher={MIT Press One Rogers Street, Cambridge, MA 02142-1209, USA journals-info~…}
}

@article{pica2004exact,
  title={Exact and approximate arithmetic in an Amazonian indigene group},
  author={Pica, Pierre and Lemer, Cathy and Izard, V{\'e}ronique and Dehaene, Stanislas},
  journal={Science},
  volume={306},
  number={5695},
  pages={499--503},
  year={2004},
  publisher={American Association for the Advancement of Science}
}

@article{o2022cultural,
  title={The cultural origins of symbolic number.},
  author={O'Shaughnessy, David M and Gibson, Edward and Piantadosi, Steven T},
  journal={Psychological review},
  volume={129},
  number={6},
  pages={1442},
  year={2022},
  publisher={American Psychological Association}
}

@book{Fritz2002,
	Address = {W\"urzburg},
	Author = {Sonja Fritz},
	Publisher = {Ergon},
	Title = {The {D}hivehi Language},
	Year = {2002}}

@article{nunez2017there,
  title={Is there really an evolved capacity for number?},
  author={N{\'u}{\~n}ez, Rafael E},
  journal={Trends in cognitive sciences},
  volume={21},
  number={6},
  pages={409--424},
  year={2017},
  publisher={Elsevier}
}

@article{dehaene1992cross,
  title={Cross-linguistic regularities in the frequency of number words},
  author={Dehaene, Stanislas and Mehler, Jacques},
  journal={Cognition},
  volume={43},
  number={1},
  pages={1--29},
  year={1992},
  publisher={Elsevier}
}

@article{baayen2018inflectional,
  title={Inflectional morphology with linear mappings},
  author={Baayen, R Harald and Chuang, Yu-Ying and Blevins, James P},
  journal={The mental lexicon},
  volume={13},
  number={2},
  pages={230--268},
  year={2018},
  publisher={John Benjamins Publishing Company Amsterdam/Philadelphia}
}

@incollection{Brysbaert2005,
	Address = {New York, Hove},
	Author = {Matthew Brysbaert},
	Booktitle = {Handbook of mathematical cognition},
	Editor = {J.I. Campbell},
	Pages = {23--42},
	Publisher = {Psychology Press},
	Title = {Number recognition in different formats},
	Year = {2005}}

@book{hurford1987language,
  title={Language and number: The emergence of a cognitive system},
  author={Hurford, James R},
  publisher={Blackwell},
  address={Oxford},
  year={1987}
}

@article{comrie2022arithmetic,
  title={The arithmetic of natural language: toward a typology of numeral systems},
  author={Comrie, Bernard},
  journal={Macrolinguistics},
  volume={10},
  number={1},
  pages={1--35},
  year={2022}
}

@inproceedings{cathcart2017decomposability,
  title={{Decomposability and Frequency in the Hindi/Urdu Number System}},
  author={Cathcart, Chundra},
  pages={1733-1738},
  booktitle={{Proceedings of the 39th Annual Meeting of the Cognitive Science Society, London}},
  year={2017},
  organization={Cognitive Science Society}
}

@article{sims-williams_token_2021,
	title = {Token frequency as a determinant of morphological change},
	doi = {10.1017/S0022226721000438},
	journal = {Journal of Linguistics},
	author = {Sims-Williams, Helem},
	year = {2021},
	note = {Publisher: Cambridge University Press},
	pages = {1--37},
}

@Article{Forkel2018a,
  author     = {Forkel, Robert and List, Johann-Mattis and Greenhill, Simon J. and Rzymski, Christoph and Bank, Sebastian and Cysouw, Michael and Hammarström, Harald and Haspelmath, Martin and Kaiping, Gereon A. and Gray, Russell D.},
  journal    = {Scientific Data},
  title      = {{C}ross-{L}inguistic {D}ata {F}ormats, advancing data sharing and re-use in comparative linguistics},
  year       = {2018},
  doi        = {https://doi.org/10.1038/sdata.2018.205},
  number     = {180205},
  pages      = {1-10},
  url        = {https://www.nature.com/articles/sdata2018205},
  volume     = {5},
  _pdf       = {https://www.nature.com/articles/sdata2018205.pdf},
  abstract   = {The amount of available digital data for the languages of the world is constantly increasing. Unfortunately, most of the digital data are provided in a large variety of formats and therefore not amenable for comparison and re-use. The Cross-Linguistic Data Formats initiative proposes new standards for two basic types of data in historical and typological language comparison (word lists, structural datasets) and a framework to incorporate more data types (e.g. parallel texts, and dictionaries). The new specification for cross-linguistic data formats comes along with a software package for validation and manipulation, a basic ontology which links to more general frameworks, and usage examples of best practices.},
  groups     = {Papers},
  keywords   = {cross-linguistic data formats, standardization, _calc, reproducibility},
  sortauthor = {List, 9},
  timestamp  = {2019-08-04},
}

@book{GnanadesikanDhivehi2017,
	Address = {Berlin, Boston},
	Author = {Amalia E. Gnanadesikan},
	Editor = {Anne E. David},
	Publisher = {De Gruyter Mouton},
	Title = {Dhivehi: The Language of the {Maldives}},
	Year = {2017}}

@article{allassonniere2020numeral,
  title={Numeral base, numeral classifier, and noun: Word order harmonization},
  author={Allassonni{\`e}re-Tang, Marc and Her, One-Soon},
  journal={Language and Linguistics},
  volume={21},
  number={4},
  pages={513--558},
  year={2020},
  publisher={John Benjamins Publishing Company Amsterdam/Philadelphia}
}

@inproceedings{beniamine2021multiple,
  title={Multiple alignments of inflectional paradigms},
  author={Beniamine, Sacha and Mat{\'\i}as {Guzm{\'a}n Naranjo}},
  booktitle={Proceedings of the Society for Computation in Linguistics 2021},
  pages={216--227},
  year={2021}
}

@incollection{Blevinsetal2017,
	Author = {Blevins, James P and Milin, Petar and Ramscar, Michael},
	Booktitle = {{Perspectives on morphological organization}},
	Pages = {139--158},
	Publisher = {Brill},
	Title = {{The Zipfian paradigm cell filling problem}},
	Year = {2017}}

\end{document}